\documentclass[11pt, a4paper]{article}

\usepackage[english]{babel}

\usepackage{authblk} 
\usepackage{hyperref} 
\usepackage{amsmath, amssymb, amsthm}
\usepackage{booktabs}
\usepackage{enumitem}
\usepackage{xcolor}
\usepackage{amsthm,amsmath,amsfonts,amssymb}
\usepackage{graphicx}
\usepackage{subcaption} 
\usepackage{todonotes}
\usepackage{stmaryrd}
\usepackage{mathtools}
\mathtoolsset{showonlyrefs}

\theoremstyle{definition}
\newtheorem{definition}{Definition}
\newtheorem{question}{Open Question}

\usepackage[style=authoryear, natbib=true, sorting=none]{biblatex}
\newcommand{\Qpre}{Q_0}
\newcommand{\Qpost}{Q_1}

\newcommand{\YY}{\boldsymbol{Y}}

\newcommand{\orcid}[1]{\href{https://orcid.org/#1}{\textcolor[HTML]{A6CE39}{\textsuperscript{ID}}}}

\title{Conformal Blindness: A Note on A-Cryptic Change-Points}

\author{Johan Hallberg Szabadv\'ary \orcid{0000-0001-7240-8919}}
\affil{Department of Computing, Jönköping School of Engineering, Jönköping \\ \texttt{johan.hallberg.szabadvary@ju.se}}

\date{\today}

\begin{document}

\maketitle

\begin{abstract}
    Conformal Test Martingales (CTMs) are a standard method within the Conformal Prediction framework for testing the crucial assumption of data exchangeability by monitoring deviations from uniformity in the p-value sequence. Although exchangeability implies uniform p-values, the converse does not hold. This raises the question of whether a significant break in exchangeability can occur, such that the p-values remain uniform, rendering CTMs blind. We answer this affirmatively, demonstrating the phenomenon of \emph{conformal blindness}. 
    Through explicit construction, for the theoretically ideal ``predictive oracle'' conformity measure (given by the true conditional density), we demonstrate the possibility of an \emph{$A$-cryptic change-point} (where $A$ refers to the conformity measure). Using bivariate Gaussian distributions, we identify a line along which a change in the marginal means does not alter the distribution of the conformity scores, thereby producing perfectly uniform p-values.
    Simulations confirm that even a massive distribution shift can be perfectly cryptic to the CTM, highlighting a fundamental limitation and emphasising the critical role of the alignment of the conformity measure with potential shifts.
    By contrasting the predictive oracle with recent results on detection-optimal scores, we emphasise that validity monitoring in safety-critical systems requires careful separation of predictive and diagnostic goals.
\end{abstract}

\section{Introduction}

    Conformal Prediction (CP) has emerged as a powerful, model-agnostic framework for predictive systems with guaranteed statistical validity \citep{alrw2}. These guarantees rest on the central assumption that the data-generating probability distribution is \emph{exchangeable}, meaning that any permutation of the data has the same probability. 
    Testing this critical assumption is crucial, and the conformal framework offers Conformal Test Martingales (CTMs) for this purpose \citep{vovk2003testing}. 

    The fundamental result, which guarantees the validity of CP, is that the p-values $p_1,p_2,\dots$ output by any conformal transducer (function that outputs conformal p-values) are independent and distributed according to the uniform distribution on the unit interval under the assumption of exchangeability. 
    CTMs operate on the natural filtration $\mathcal{F}_n = \sigma(p_1,\dots,p_n)$ generated by the p-values by betting against their uniformity. This ``impoverished'' filtration contains less information than the canonical filtration generated by the sequence of examples $z_1,z_2,\dots$, but the canonical filtration (generated by $z_i$) allows only trivial (constant) martingales, while the impoverished filtration lets us construct several non-trivial martingales \citep{ramdas2022testing}. If the CTM grows large, this is direct evidence that the exchangeability assumption is violated (see Ville's theorem \citep{ville1939etude}).

    However, the implication goes in only one direction. Exchangeability implies uniform p-values but not vice versa. This raises concerns: is it possible to have a major violation of exchangeability, for example, a drastic change-point in the data-generating distribution, in such a way that the p-values remain uniform after the shift? If so, any CTM would be completely blind to such a violation, rendering the test impotent and leaving predictive systems vulnerable to undetected distribution shifts.

    To understand the limits of detectability, it is important to distinguish the ``Predictive Oracle'' (the true conditional density) from the ``Detection Oracle'' (the likelihood ratio). While the latter has been shown to be asymptotically consistent for detecting exchangeable shifts \cite{bhattacharyya2025theoretical, dandapanthula2025conformal}, practitioners often rely on the former for monitoring because it minimises the prediction error. We demonstrate that this reliance on predictive optimality creates structural vulnerabilities, $A$-cryptic pairs, that do not exist for detection-optimal scores.

    In this note, we demonstrate that such \emph{conformal blindness} is indeed possible, despite the use of the best possible conformal transducer, which is based on full knowledge of the pre-change distribution. We construct what we call an \emph{$A$-cryptic change-point} (where $A$ refers to the conformity measure defined below), which is impossible to detect using any CTM because the p-values remain uniform. This finding highlights a fundamental limitation of conformal testing and serves as a cautionary tale regarding its application.

\section{Problem formulation}

    At the heart of standard Conformal Prediction (CP) lies the assumption of \emph{exchangeability}. A sequence of random variables $(Z_1, Z_2, ...)$ is exchangeable if its joint probability distribution is invariant under any permutation of the indices \citep{alrw2, bernardo1996concept}. This condition, which is weaker than assuming independent and identically distributed (IID) data, forms the basis for CP's validity guarantees.

    CP procedures typically employ a \emph{nonconformity measure}, denoted $A$, to quantify how ``strange'' or ``unusual'' an example $z_i$, $i=1,\dots,n$ appears relative to a bag (or multiset) of observed examples $\lbag z_1, ..., z_n\rbag$. Higher scores indicate greater nonconformity. It is also possible to use a \emph{conformity measure}, whose purpose is the reversed; score how ``typical'' an example is. We use the latter in our construction.
    When observing a test object $x_n$, we form a tentative new example $z_n^y:=(x_n,y)$ and compute its \emph{nonconformity score} $\alpha_n$ using $A$. 
    This score is then used to compute a \emph{p-value}, defined as the fraction of examples (including the tentative new example) whose nonconformity scores are greater than or equal to the score of the new example:
    \begin{equation}
        p_n^y := \frac{|\{i=1,\dots,n:\alpha_i\geq\alpha_n\}|}{n},
    \end{equation}
    where 
    \begin{equation}
        \begin{aligned}
            \alpha_i &= A(\lbag z_1, ..., z_{n-1},(x_n,y)\rbag, z_i), ~i=1,\dots,n-1\\
            \alpha_n &= A(\lbag z_1, ..., z_{n-1},(x_n,y)\rbag, (x_n, y)).
        \end{aligned}
    \end{equation}
    When using a conformity score, replace ``greater than'' with ``less than''. For brevity, in this note, we will use a \emph{conformity measure}, but the discussion applies equally to nonconformity measures.
    \emph{Smoothed p-values} are constructed by using auxiliary randomness to break ties in the p-value calculation.
    
    This process is repeated for each candidate label $y$, resulting in $|\YY|$ p-values associated with the test object $x_n$, one for each potential label. The conformal prediction set is formed by including all labels whose p-values exceed a user-specified significance level $\varepsilon$. 
    Intuitively, the p-value $p_y$ measures the ``typicality'' of the new example $z_n^y$ relative to the existing data, as judged by the conformity measure $A$. The conformal prediction set is all candidate labels $y$ that appear ``typical enough''.

    It is sometimes convenient to define a function $f$ that transforms a sequence of examples $z_1,z_2,\dots$ into a sequence of p-values $p_1,p_2,\dots$. This function is called the \emph{conformal transducer} determined by the conformity measure $A$. A conformal transducer that outputs smoothed p-values is called a smoothed conformal transducer. The main result in CP is that if data are generated by an exchangeable probability distribution, the p-values output by any smoothed conformal transducer (denoted $p_1,p_2,\dots$) are IID uniform on the unit interval. CTMs rely on this result; they are designed to detect deviations from p-value uniformity, thereby providing a test for the underlying exchangeability assumption \citep{vovk2003testing}. In conformal testing, only smoothed p-values are considered because they are what the CTMs tests. Therefore, by ``p-value'' we always mean ``smoothed p-value'' in this note.

    Importantly, exchangeability is sufficient for uniform p-values, but not necessary. This raises the possibility of a distribution shift occurring in a manner that leaves the p-value stream undisturbed.

    \begin{definition}\label{def:A-cryptic pair}
        A pair of probability distributions $(\Qpre, \Qpost)$ is called $A$-cryptic, where $A$ is a conformity measure if the conformal transducer determined by $A$ continues to produce IID uniform p-values if the data-generating distribution changes from $\Qpre$ to $\Qpost$.
    \end{definition}
    A change-point that shifts from $\Qpre$ to $\Qpost$, where  $(\Qpre, \Qpost)$ is $A$-cryptic, is called an \emph{$A$-cryptic change-point}.
    Note that a sufficient condition for $(\Qpre, \Qpost)$ to be $A$-cryptic is that the distributions of conformity scores under $\Qpre$ and $\Qpost$ are identical. It is unclear a priori whether $A$-cryptic pairs exist, but they are not theoretically forbidden. The next section proves by construction that $A$-cryptic pairs exist in the case of bivariate Gaussian distributions using the predictive oracle (or idealised) conformity measure, which is given by the conditional probability density function of $Y$. Not only can we find such pairs, but we can also identify a line in $\mathbb{R}^2$ along which a change in mean is $A$-cryptic.
    
\section{Construction}
    Let $Q_0$ be a bivariate Gaussian distribution with mean $\mu_0 = (\mu_{0X},\mu_{0Y})^T$ and covariance matrix 
    \begin{equation}
        \Sigma_0 = 
        \begin{pmatrix}
            \sigma^2_{0X} & \rho_0 \\
            \rho_0 & \sigma^2_{0Y}
        \end{pmatrix}.
    \end{equation}
    As conformity measure we use the conditional probability density function
    \begin{equation}
        A(x, y) := f_{Y\mid X}(y\mid x).
    \end{equation}
    We deliberately choose this \emph{"predictive oracle" conformity measure}, $A(x,y) := f_{Y|X}(y|x)$, which assumes knowledge of the true conditional density under $Q_0$. The primary goal of an ideal conformity measure is to produce the most efficient (i.e. smallest or "tightest") prediction set, whereas the CP framework itself guarantees validity. Efficiency is achieved by a maximally discriminating conformity measure: it should assign the highest possible conformity score to the true, unobserved label $y_{\text{true}}$ and low scores to all incorrect labels.

    By definition, the true conditional density $f_{Y|X}(y|x)$ is the function that perfectly captures the "typicality" of $y$ given $x$, assigning the highest likelihood precisely to the most probable labels. This choice represents the "soft model" approach discussed in \citet{alrw2}: we leverage a probabilistic model (the predictive oracle $f_{Y|X}$) as the engine for our score to provide Bayesian-like efficiency, while using the CP framework to provide a frequentist coverage guarantee.

    Therefore, choosing $A = f_{Y|X}$ is not arbitrary; it represents the theoretical optimum for prediction accuracy.

    The parameters of the conditional PDF are
    \begin{equation}
        \begin{aligned}
            \mathbb{E}[Y\mid X=x] &= \mu_{0Y} + \rho_0\sigma_{0Y}\frac{x - \mu_{0X}}{\sigma_{0X}}\\
            \text{Var}[Y\mid X=x] &= (1-\rho_0^2)\sigma_{0Y}^2
        \end{aligned}            
    \end{equation}
    Now, the question is whether we can construct another distribution $Q_1$ such that the conformal transducer determined by $A$ outputs uniform p-values after shifting from $Q_0$ to $Q_1$. 

    \subsection*{Sufficient conditions for a conformal-cryptic change}
        We aim to construct $Q_1$ such that the distribution of conformity scores remains identical. This suffices to make the change invisible to a conformal test martingale since the p-values remain uniform. This gives us two conditions.
        \begin{enumerate}
            \item 
            \textbf{Conditional mean invariance}: The conditional mean under $Q_1$ must be the same as under $Q_0$.
            $$
                \mu_{0Y} + \rho_0\sigma_{0Y}\frac{x - \mu_{0X}}{\sigma_{0X}} = \mu_{1Y} + \rho_1\sigma_{1Y}\frac{x - \mu_{1X}}{\sigma_{1X}}
            $$
            \item 
            \textbf{Conditional variance invariance}: The conditional variance under $Q_1$ must be the same as under $Q_0$.
            $$
                (1-\rho_0^2)\sigma_{0Y}^2 = (1-\rho_1^2)\sigma_{1Y}^2
            $$
        \end{enumerate}
        Assuming that the covariance matrix remains fixed, that is, $\Sigma_0=\Sigma_1=:\Sigma$ ensures that Condition 2 is satisfied because $\sigma^2_{0X}= \sigma^2_{1X} =: \sigma^2_X, \sigma^2_{0Y} = \sigma^2_{1Y}=:\sigma^2_Y$ and $\rho_0=\rho_1=:\rho$. Then Condition 1 simplifies to
        \begin{equation}\label{eq:condition1Simplified}
            \mu_{0Y} + \rho\sigma_{Y}\frac{x - \mu_{0X}}{\sigma_{X}} = \mu_{1Y} + \rho\sigma_{Y}\frac{x - \mu_{1X}}{\sigma_{X}}.
        \end{equation}
        For \eqref{eq:condition1Simplified} to hold for all $x\in\mathbb{R}$, we must have   
        $$
            \mu_{0Y} - \rho\sigma_{Y}\frac{\mu_{0X}}{\sigma_{X}} = \mu_{1Y} - \rho\sigma_{Y}\frac{\mu_{1X}}{\sigma_{X}}.
        $$
        Equivalently,
        \begin{equation}\label{eq:condition1SimplifiedFull}
            \mu_{1Y} - \mu_{0Y} = \rho\frac{\sigma_Y}{\sigma_X}(\mu_{1X} - \mu_{0X}),
        \end{equation}
        showing that we can change the mean of $X$ and $Y$ as long as \eqref{eq:condition1SimplifiedFull} holds; that is, the change in $\mu_Y$ must be proportional to the change in $\mu_X$ by a factor of $\rho\frac{\sigma_Y}{\sigma_X}$. This implies that the mean can change to any point along the line 
        $$
            y = \mu_{0Y} + \rho\frac{\sigma_Y}{\sigma_X}(x - \mu_{0X}).
        $$ 
        We found an entire line of $A$-cryptic shifts.

\section{Illustration}\label{sec:construction}
    Imagine a forecaster with a strong belief that the data-generating process is a bivariate Gaussian distribution with known parameters. Her goal is to predict the label $y$ with high confidence. This can be done directly from the conditional density function, but she wants to ensure that her predictions are valid, even if her belief is wrong. She can then use the Gaussian model as a ``soft model'', as discussed in \citet{alrw2}. Essentially, she employs the CP framework as an insurance policy, should her belief be wrong. She uses the predictive oracle conformity measure and monitors the p-values as she goes with a CTM.

\begin{figure}[htbp]
            \centering
            
            \begin{subfigure}[b]{0.8\textwidth}
                \centering
                \includegraphics[width=\textwidth]{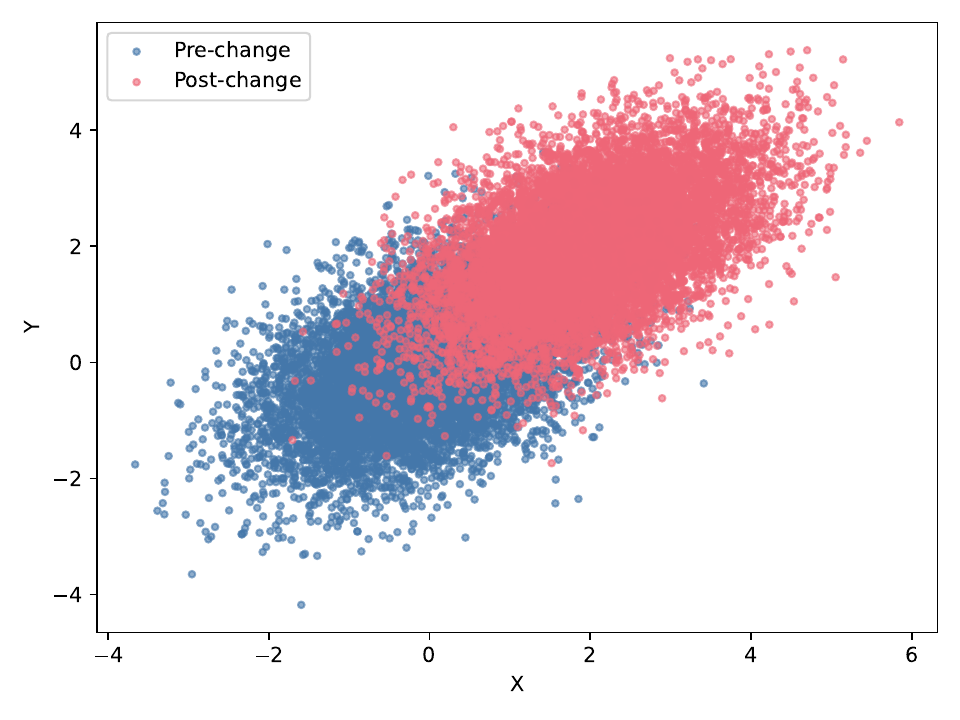} 
                \caption{Non-Cryptic Data Shift}
                \label{fig:scatter_noncryptic}
            \end{subfigure}
            
            \vspace{1cm} 
        
            \begin{subfigure}[b]{0.48\textwidth}
                \centering
                \includegraphics[width=\textwidth]{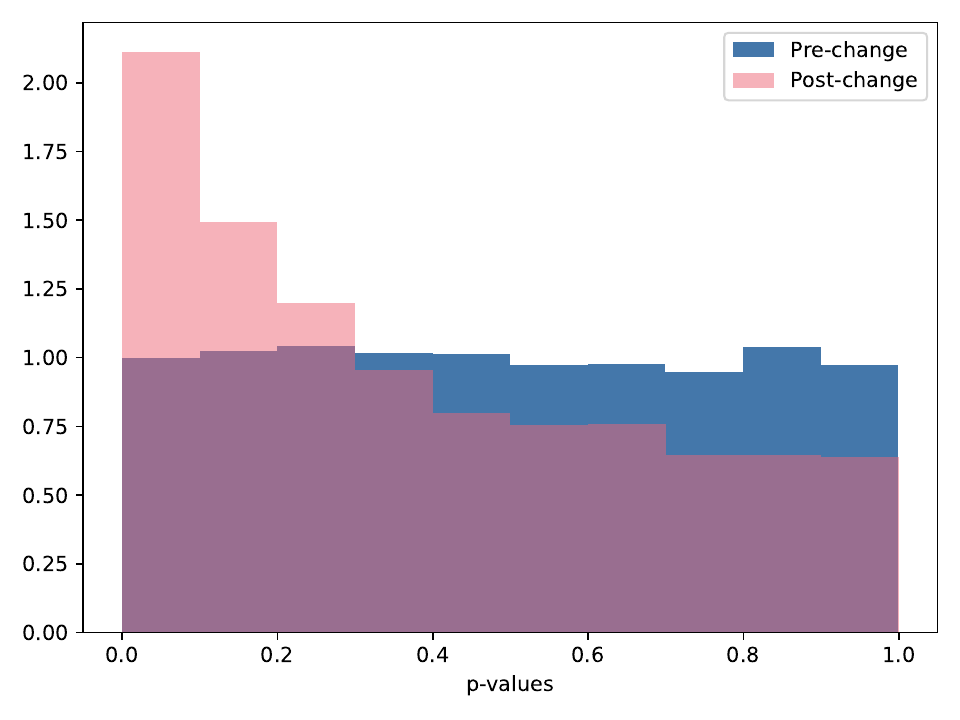}
                \caption{Skewed p-value Histogram}
                \label{fig:hist_noncryptic}
            \end{subfigure}
            \hfill 
            \begin{subfigure}[b]{0.48\textwidth}
                \centering
                \includegraphics[width=\textwidth]{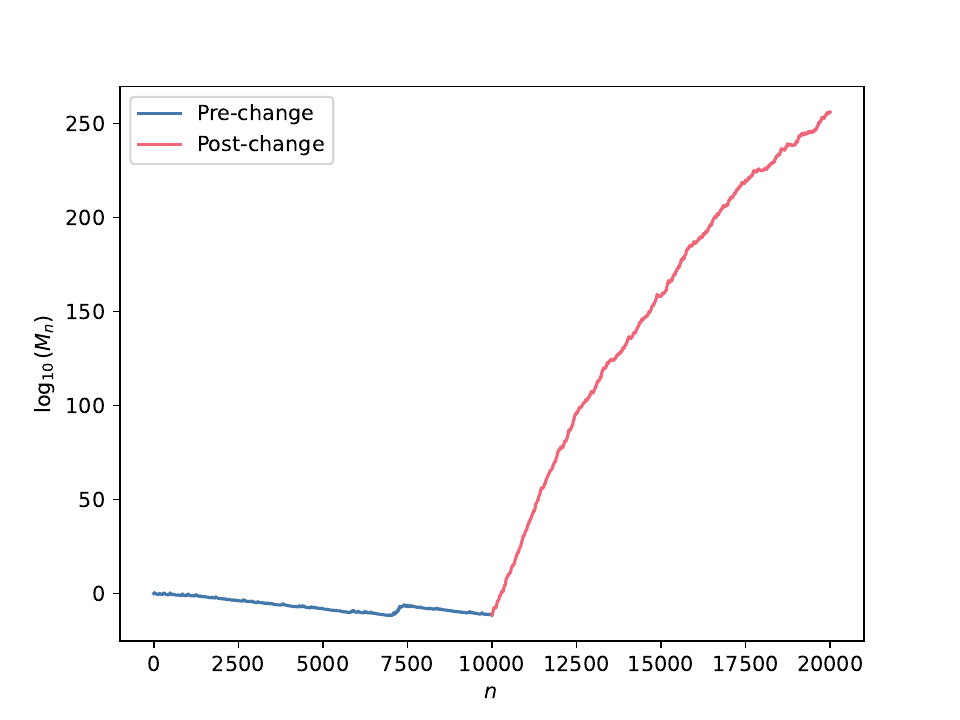}
                \caption{Successful CTM Growth}
                \label{fig:ctm_noncryptic}
            \end{subfigure}
            
            \caption{Detection of a non-cryptic change-point. 
            (a) A shift in data distribution occurs. 
            (b) The resulting p-value distribution becomes clearly non-uniform. 
            (c) Consequently, the Conformal Test Martingale successfully grows, detecting the break in exchangeability.}
            \label{fig:non-cryptic}
        \end{figure}

    Figure \ref{fig:non-cryptic} illustrates a typical change-point situation that is not $A$-cryptic. We generate 10000 examples drawn from $\Qpre$, a multivariate Gaussian distribution with mean $\mu_0 = (0,0)^T$ and covariance matrix
        \begin{equation}
            \Sigma = 
            \begin{pmatrix}
            \sigma^2_{X} & \rho \\
            \rho & \sigma^2_{Y}
        \end{pmatrix}
        \end{equation}
        with $\sigma^2_X=\sigma^2_Y = 1$ and $\rho=1/2$
        followed by 10000 examples from a multivariate Gaussian distribution $\Qpost$ with $\mu_1 = (2,2)^T$ and the same covariance matrix. Note that $\mu_1$ does not lie on the $A$-cryptic line. 
        Panel (a) shows a scatter plot of the pre- and post-change realised values.
        The pre- and post-change histograms of the p-values, shown in panel (b), clearly show a post-change p-value distribution that is skewed and far from uniform.
        The exchangeability assumption is tested by the Simple Jumper martingale introduced in \citet{vovk2021retrain}. Before the change-point (blue trajectory in panel (c)), it remains at low values; however, once the change occurs, it quickly explodes to enormous values. The final value is larger than $10^{255}$, which we can contrast with the ``mere'' $10^{80}$ estimated atoms in the observable universe. The test firmly rejects exchangeability which is expected because we used the predictive oracle conformity measure $A$ to detect the change. 

\begin{figure}[htbp]
        \centering
        
        \begin{subfigure}[b]{0.8\textwidth}
            \centering
            \includegraphics[width=\textwidth]{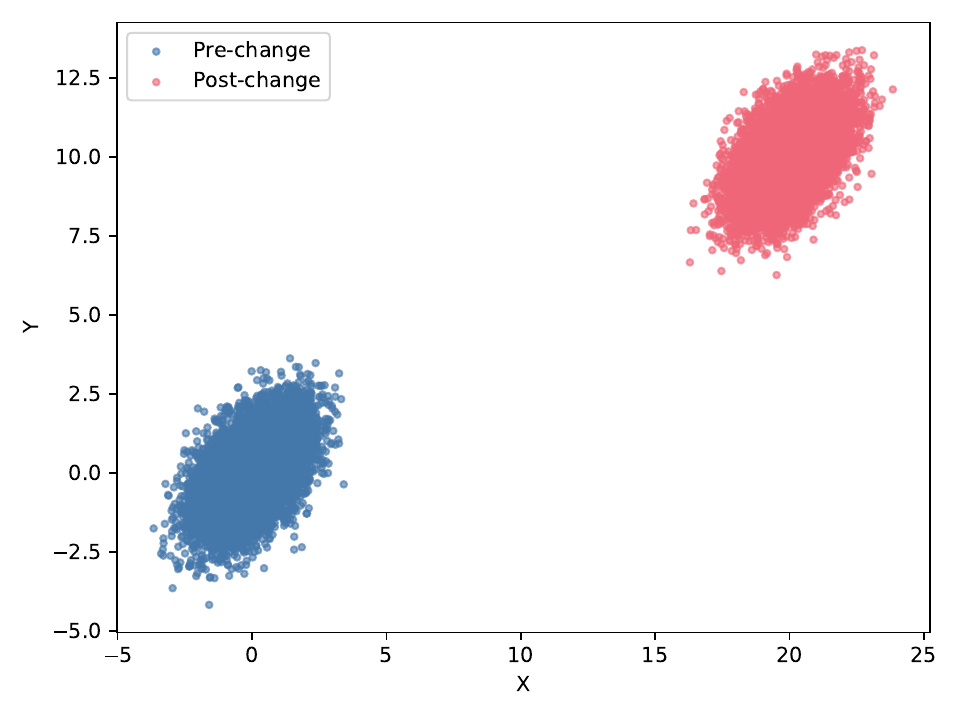}
            \caption{$A$-cryptic data shift}
            \label{fig:scatter}
        \end{subfigure}
        
        \vspace{1cm} 
    
        \begin{subfigure}[b]{0.48\textwidth}
            \centering
            \includegraphics[width=\textwidth]{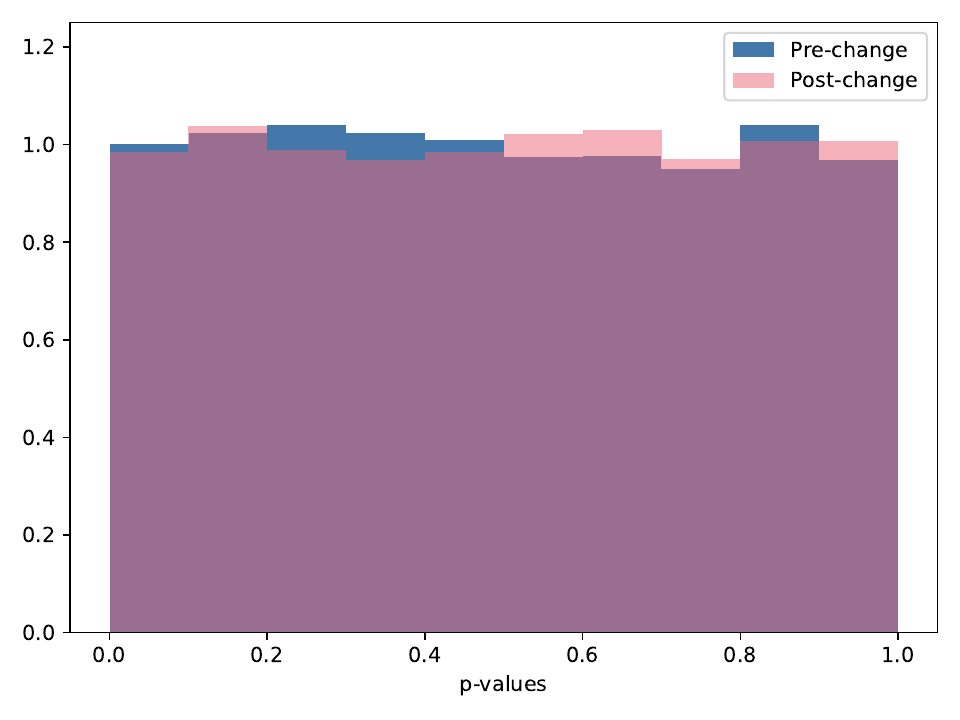}
            \caption{p-value Histogram}
            \label{fig:hist}
        \end{subfigure}
        \hfill 
        \begin{subfigure}[b]{0.48\textwidth}
            \centering
            \includegraphics[width=\textwidth]{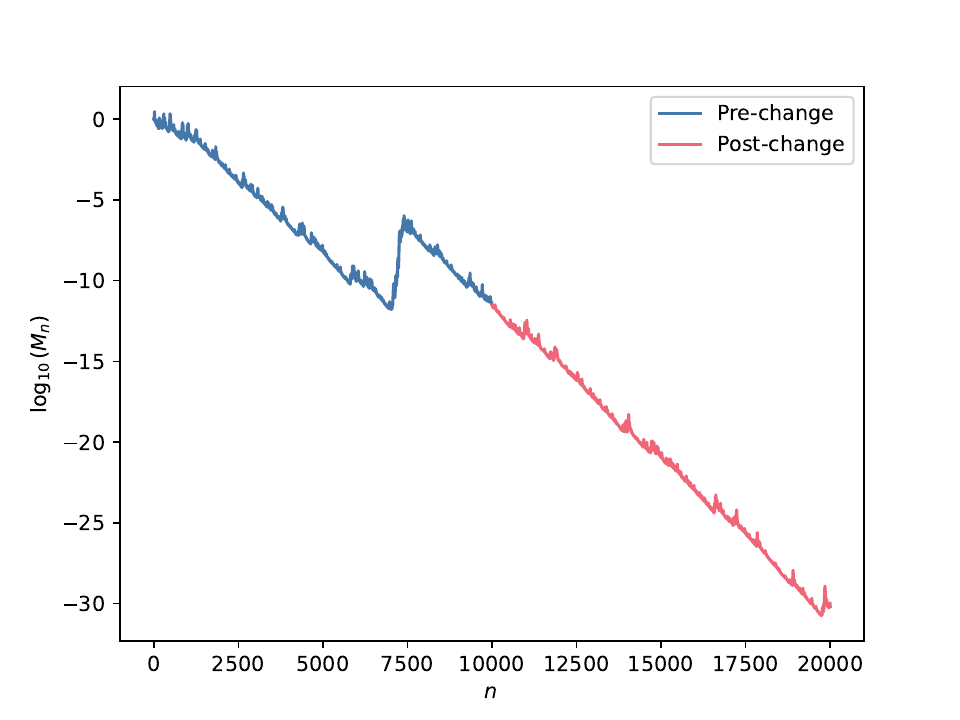}
            \caption{CTM Trajectory}
            \label{fig:ctm}
        \end{subfigure}
        
        \caption{Visual demonstration of a conformal-cryptic change-point. 
        (a) A massive shift in the data distribution is visually apparent. 
        (b) The resulting p-value histogram remains perfectly uniform. 
        (c) Consequently, the Conformal Test Martingale fails to grow, remaining blind to the shift.}
        \label{fig:cryptic}
    \end{figure}

    In contrast, Figure \ref{fig:cryptic} shows a larger shift in the marginal means. Now, $\Qpost$ has mean $\mu_1 = (20,10)^T$ (but still the same covariance matrix). This shift lies along the $A$-cryptic line. The scatter plot in panel (a) illustrates that this change-point results in two completely separate clusters. The change is very dramatic and should be easy to detect, especially because we use oracle knowledge.
    Nevertheless, the p-value histograms in panel (b) are almost perfectly flat and the martingale trajectory in panel (c) decreases and does not increase, as expected. The predictive oracle conformity measure is indeed blind to this drastic change.
    
\begin{figure}[htbp]
        \centering
        
        \begin{subfigure}[b]{0.48\textwidth}
            \centering
            \includegraphics[width=\textwidth]{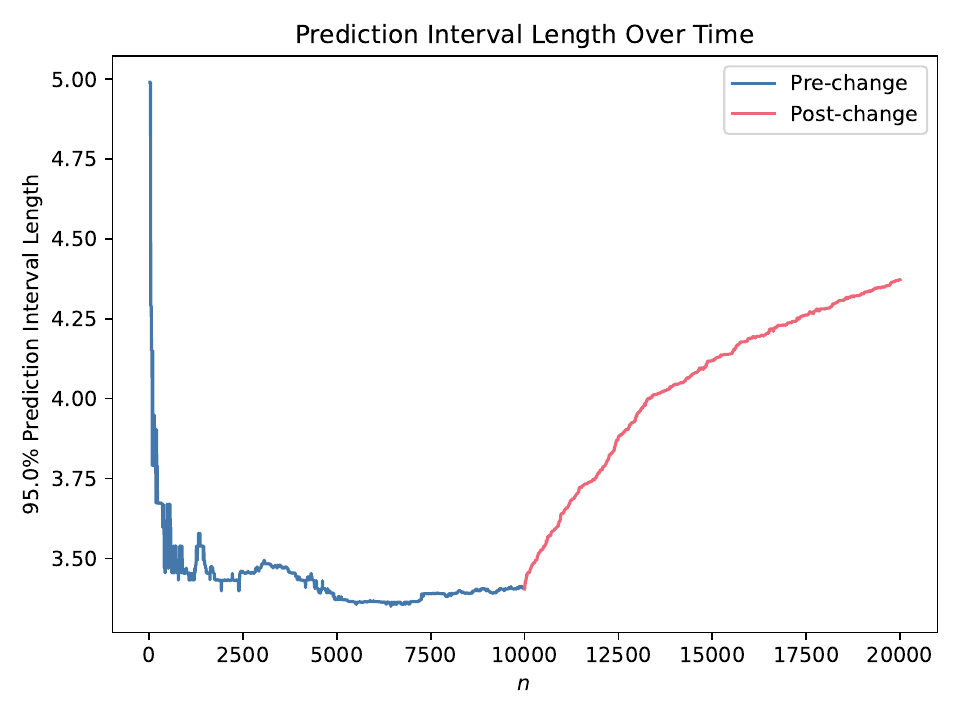}
            \caption{Non-cryptic data shift}
            \label{fig:effNonCrypt}
        \end{subfigure}
        \hfill
        \begin{subfigure}[b]{0.48\textwidth}
            \centering
            \includegraphics[width=\textwidth]{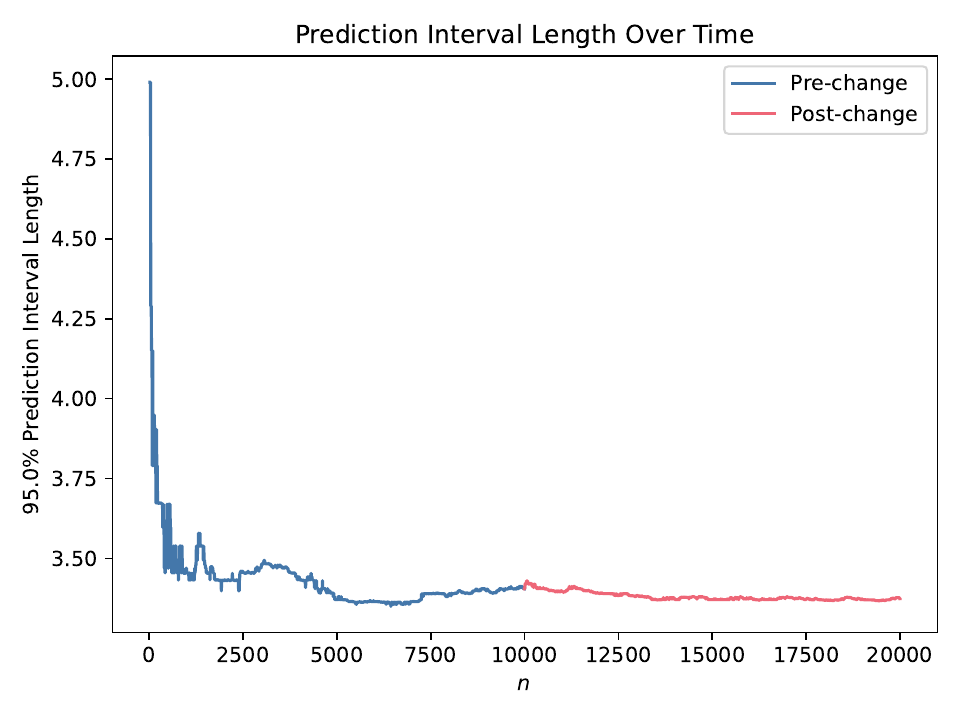}
            \caption{$A$-cryptic data shift}
            \label{fig:effCrypt}
        \end{subfigure}
        
        \caption{Visual demonstration of predictive efficiency. 
        (a) A moderate shift in the data distribution results in clearly increasing interval lengths. 
        (b) A massive $A$-cryptic shift results in no change in efficiency. Prediction intervals remain valid \emph{and} efficient.}
        \label{fig:efficiency}
    \end{figure}

    Going back to our forecaster, in both cases, her belief turned out to be correct for the pre-change data. In the first case, she was able to detect when the change occurred and (one must assume) update her beliefs accordingly. In the second case, her belief was suddenly violated drastically, but she was unable to notice this. This is not a failure of the Simple Jumper martingale. No CTM could have detected the change, because the p-values remained uniform. Uniform post-change p-values mean that despite her belief about the data-generating process being wrong, the forecaster's predictions are still valid. Interestingly, they also suffer no loss of efficiency (informally, ``tightness''). As illustrated in Figure \ref{fig:efficiency}, the non-cryptic change (Figure \ref{fig:effNonCrypt}) results in invalid intervals that are clearly wider than before the shift. In contrast, no loss of efficiency is apparent in Figure \ref{fig:effCrypt}. The intervals remain, on average, the same width. 

    As far as our forecaster is concerned, her belief is wrong, but only in those respects that are irrelevant to her predictions.
    
\section{Discussion}
    Strengthening the reliability of predictive systems with statistical guarantees, such as those offered by CP, requires a deep understanding of their potential limitations. In this spirit, we demonstrated through explicit construction the existence of $A$-cryptic change-points. Our construction shows that the predictive oracle conformity measure is blind to shifts in the marginal means because, by definition, it is sensitive only to the conditional distribution of $Y$ given $X$. The A-cryptic line is precisely the set of mean shifts that leave this conditional distribution invariant. Figure \ref{fig:cryptic} provides a striking visual confirmation: despite a drastic change in the mean, the p-value histograms remain perfectly uniform, and the CTM fails to detect the change. This contrasts sharply with Figure \ref{fig:non-cryptic}, where a non-cryptic shift correctly triggers the CTM.

    However, this blindness reveals a fascinating paradox that is worth exploring. As shown in Section \ref{sec:construction}, the $A$-cryptic shift leaves the p-value distribution unchanged. Consequently, the forecaster's predictions are still valid and, as Figure \ref{fig:effCrypt} illustrates, suffer no loss of efficiency. The very reason the CTM fails (p-value uniformity) is the exact same reason that the CP guarantees are upheld. This implies that CTMs do not test for all breaks in exchangeability, but rather for a specific subset: those shifts that are \emph{$A$-harmful}, that is, those that actually disrupt the p-value uniformity and, by extension, the predictive guarantees. The existence of $A$-cryptic shifts suggests that the CP framework itself can be robust to some forms of distribution shifts, even if the CTM is blind to them.

    This finding highlights a fundamental limitation in conformal testing. CTMs are theoretically grounded, but their ability to detect deviations from exchangeability is entirely dependent on the p-values reflecting the deviation. If they remain uniform, betting against their uniformity fails. Our work shows that this reflection is not guaranteed, even when using an ideal predictive oracle conformity measure based on true conditional density.
    
    This directly leads to a crucial open question for future investigation.
    \begin{question}
        Under what conditions on the conformity measure $A$ do $A$-cryptic pairs $(\Qpre,\Qpost)$ exist?
    \end{question}
    Our Gaussian example provides a specific instance; however, a general characterisation is lacking. Answering this question is vital. It is possible that for most practical conformity measures, such as those based on flexible models (e.g. random forest or neural networks), perfect cryptic shifts are rare or even impossible. 
    
    This phenomenon highlights a potential vulnerability to adversarial attacks, in which an adversary who knows $A$ and $\Qpre$ may be able to design a shift such that $(\Qpre,\Qpost)$ forms an $A$-cryptic pair to evade detection. This identifies a more specific version of the open question.
    \begin{question}
        Under what conditions on the conformity measure $A$ and a pre-change distribution $\Qpre$ can a post-change distribution $\Qpost$ be found such that $(\Qpre,\Qpost)$ is $A$-cryptic?
    \end{question}
    Our construction also underscores that the choice of $A$ implicitly defines the types of shift that a CTM can monitor. In the specific Gaussian case discussed in this note, the Mahalanobis distance serves as a simple example of a measure that breaks the crypticity of predictive scores. Theoretically, the most powerful alternative for detection is the Likelihood Ratio (LR) score, as formalised by the Conformal Neyman-Pearson lemma \cite{dandapanthula2025conformal}. 

    However, our results regarding $A$-cryptic pairs remain a cautionary tale: even the most efficient model for prediction may be perfectly blind to semantic inversions if those inversions align with the model's level sets. This underscores that while optimal detection scores exist in theory, relying on standard predictive models for safety monitoring introduces structural blind spots that are not easily mitigated without explicit knowledge of potential shifts.

    Given this potential for cryptic changes, a practitioner may employ an ensemble of conformity measures, using, for example, their convex combination as the final conformity measure. In our example, the convex combination of the predictive oracle conformity measure and one based on the Mahalanobis distance detects the change (the Mahalanobis distance is sensitive to shifts in the unconditional means, which is precisely what our A-cryptic shift alters). However, without a general characterisation of the condition for $A$-cryptic pairs, it may still be blind to other changes.

    Another potential remedy for practitioners seeking robustness might be to employ several conformal transducers to produce several sequences of p-values. Each of these could be tested separately. However, this introduces statistical inefficiencies.
    To manage multiple resulting tests and control for false positives, a method such as Bonferroni correction \citep{bonferroni1936teoria} is required. This makes the detection threshold for each test stricter, potentially reducing the ability to detect subtle changes. 
    
    A more sophisticated alternative involves monitoring the joint distribution of the p-value vectors ($p_i := (p_{i,1},\dots,p_{i,k})$ for $k>1$ transducers). Although this approach is more powerful, it is also more complex because this joint distribution, which, under exchangeability, has uniform marginals and is thus a copula, is not itself uniform. The development of such joint tests remains a direction for future research (e.g. \citet{bostrom2025testing}).

    Conformal transducers operate in two distinct modes: online or \emph{transductive}, where conformity scores are computed using all available data for both training and calibration, and \emph{inductive} (or split), where a training set is partitioned into a \emph{proper training set} used to train a conformity measure and a \emph{calibration set} used to compute p-values. Their validity is strongest in the online mode but comes at a high computational cost. For this reason, inductive CP, which operates offline, was suggested as a computationally efficient version of CP. 
    The difference relevant to our purpose is that an inductive conformal transducer, once trained, is a fixed function. 
    Our predictive oracle measure, which is a fixed function (i.e. true conditional density), mimics an inductive conformal transducer. Therefore, it serves as a direct analogue of an inductive conformal transducer. This, at least intuitively, suggests that inductive conformity measures may support $A$-cryptic pairs. 
    Transductive conformal transducers, however, are constantly updated as new data arrive, which suggests that they may be more robust to conformal blindness. The reasoning is that the bag of examples used for calibration continually incorporates post-change data, progressively altering the conformity scores and potentially disrupting the conditions for a cryptic shift. However, this hypothesis remains speculative.

    Importantly, our findings do not invalidate the practical utility of CTMs. Rather, they underscore the critical importance of the conformity measure's role, not just in prediction but also in detection. The choice of $A$ implicitly defines the types of distribution shifts that a CTM can monitor effectively. The conditional density conformity measure is ideal for prediction accuracy but is apparently blind to changes in the marginal means along a line.

    In conclusion, this note demonstrates the phenomenon of conformal blindness. While CTMs remain a valuable tool, our findings reveal a fundamental limitation tied to the choice of conformity measure, $A$. The central takeaway is the need for a general characterisation of the conditions that create these A-cryptic blind spots.

\section*{Acknowledgements}
    The author thanks Vladimir Vovk for useful discussions and Tuwe Löfström for suggesting the adversarial attack angle. The author is also grateful to Aaditya Ramdas for helpful feedback and for pointing out the relevance of the Likelihood Ratio oracle. 
    Johan Hallberg Szabadv\'ary is also affiliated with the Department of Mathematics, Stockholm University, Stockholm, Sweden.

\printbibliography

\end{document}